# FCM Based Blood Vessel Segmentation Method for Retinal Images

[1]Nilanjan Dey, [2]Anamitra Bardhan Roy, [3]Moumita Pal, [4]Achintya Das

[1]Asst. Prof., Dept. of Information and Technology, JIS College of Engineering
Kalyani, West Bengal, India

[2]BTech. Student, Dept. of Computer Science and Engineering, JIS College of Engineering
Kalyani, West Bengal, India

[3]Asst. Prof., Dept. of Elec. and Comm. Engineering, JIS College of Engineering
Kalyani, West Bengal, India

[4]Professor and Head,Elec. and Telecom  Engg Dept.Kalyani Govt. Engineering College.
Kalyani, West Bengal, India.

## Abstract

Segmentation of blood vessels in retinal images provides early diagnosis of diseases like glaucoma, diabetic retinopathy and macular degeneration. Among these diseases occurrence of Glaucoma is most frequent and has serious ocular consequences that can even lead to blindness, if it is not detected early. The clinical criteria for the diagnosis of glaucoma include intraocular pressure measurement, optic nerve head evaluation, retinal nerve fiber layer and visual field defects. This form of blood vessel segmentation helps in early detection for ophthalmic diseases, and potentially reduces the risk of blindness.

The low-contrast images at the retina owing to narrow blood vessels of the retina are difficult to extract. These low contrast images are, however useful in revealing certain systemic diseases. Motivated by the goals of improving detection of such vessels, this present work proposes an algorithm for segmentation of blood vessels, and compares the results between expert ophthalmologists' hand-drawn ground-truths and segmented image (i.e. the output of the present work). Sensitivity, specificity, positive predictive value (PPV), positive likelihood ratio (PLR) and accuracy are used to evaluate overall performance. It is found that this work segments blood vessels successfully with sensitivity, specificity, PPV, PLR and accuracy of 99.62%, 54.66%, 95.08%, 219.72 and 95.03%, respectively.

## 1. Introduction

Current methods of detection and assessment of diabetic retinopathy [4] are manual, expensive and require trained ophthalmologists. Retinal blood vessel [7] morphology can be an important indicator for many diseases such as diabetes, hypertension and arteriosclerosis. The measurement of geometrical changes in veins and arteries can be applied to a variety of clinical studies. Two major problems in the segmentation of retinal blood vessels are the presence of a wide variety of vessel widths and the heterogeneous background of the retina. Retinal images provide considerable information on pathological changes caused by local ocular diseases revealing diabetes, hypertension, arteriosclerosis, cardiovascular disease and stroke. Computer-aided analysis of retinal image plays a central role in diagnostic procedures. However, automatic retinal segmentation is complicated by the fact that retinal images are often noisy, poorly contrasted, and the vessel widths can vary from very large to very small value. For this specific reason, in this work the preprocessing step includes adaptive thresholding and contrast enhancement. Segmentation of blood vessels has been a research area, for years. This present work proposes algorithms that usually use some kind of vessel enhancement before thresholding or vessel tracking. The methods with high accuracy also have high computational needs, if thick vessels are present.

The use of the proposed resolution hierarchy makes it possible to detect these vessels faster, while preserving a high accuracy.

There are three basic approaches for automated segmentation of blood vessels [8]: thresholding method, tracking method and machine trained classifiers. In the first method, many different operators are used to enhance the contrast between vessel and background, such as Sobel operators, Laplacian operators, Gaussian filters modeling the gray cross-section of blood vessel. Then the gray threshold is selected to determine the vessel. This gray threshold is crucial, because small threshold induces more noise and gray threshold causes loss of some fine vessels, adaptive or local threshold is used. Vessel tracking is another technique for vessel segmentation, whereby vessel centre locations are automatically sought along the vessel longitudinal axis from a starting point to the ending point.

The Fuzzy C-Means (FCM) clustering is a well-known clustering technique for image segmentation. It was developed by Dunn and improved by Bezdek. It has also been used in retinal image segmentation. Osareh *et al.* used color normalization and a local contrast enhancement in a pre-processing step.

## 2. Methodology

### 2.1 Fuzzy C-Means (FCM)

In pattern recognition a clustering method known as Fuzzy C-Means (FCM) is widely used. FCM [2], proposed by Bezdek in 1973[6], is also known as Fuzzy ISODATA [5]. FCM based segmentation is fuzzy pixel classification. In this clustering technique one piece of data belongs to two or more clusters. FCM allows data points or pixels to belong to multiple classes with varying degree of membership function between 0 to 1.

FCM possesses unique advantage of grading linguistic variables to fit for appropriateate analysis in discrete domain on pro-rata basis.

FCM[1] computes cluster centres or centroids by minimizing the dissimilarity function with the help of iterative approach. By updating the cluster centres and the membership grades for individual pixel, FCM shifts the cluster centres to the "right" location within set of pixels.

To accommodate the introduction of fuzzy partitioning, the membership matrix (U) =[$u_{ij}$] is randomly initialized according to Equation 1, where $u_{ij}$ being the degree of membership function of the data point of $i^{th}$ cluster $x_i$.

$$\sum_{i=1}^{c} u_{ij} = 1, \forall j = 1,...,n \quad (1)$$

The performance index (PI) for membership matrix U and $C_i$'s used in FCM is given Equation 2.

$$J(U, c_1, c_2,...,c_c) = \sum_{i=1}^{c} J_i = \sum_{i=1}^{c}\sum_{j=1}^{n} u_{ij}^{m} d_{ij}^{2} \quad (2)$$

$u_{ij}$    is between 0 and 1.
$c_i$     is the centroid of cluster i.
$d_{ij}$   is the Euclidian distance between $i^{th}$ centroid ($c_i$) and $j^{th}$ data point.
m ϵ [1,∞] is a weighting exponent.

To reach a minimum of dissimilarity function there are two conditions. These are given in Equation 3 and Equation 4.

$$c_i = \frac{\sum_{j=1}^{n} u_{ij}^{m} x_j}{\sum_{j=1}^{n} u_{ij}^{m}} \quad (3)$$

$$u_{ij} = \frac{1}{\sum_{k=1}^{c} \left(\frac{d_{ij}}{d_{kj}}\right)^{2/(m-1)}} \quad (4)$$

**Algorithm of FCM**

Step1.   The membership matrix (U) that has constraints in Eqn 1 is randomly initialized.
Step2.   Centroids($c_i$) are calculated by using Eqn 3.
Step3.   Dissimilarity between centroids and data points using Eqn 2 is computed. Stop if its improvement over previous iteration is below a threshold.
Step4.   A new U using Eqn 4 is computed. Go to Step 2.

## 3. Proposed Method

Step 1. The Color Retinal Fundus Image in Gray scale is converted from green channel.
Step 2. Adaptive histogram equalization [6] is carried out on the gray image.
Step 3. The background is subtracted from the foreground of the image using median filter.
Step 4. FCM is applied on the image followed by binarization and filtering.
Step 5: The ground truth image is compared with the corresponding disease.
Step 6. The Sensitivity, Specificity, PPV, PLR and Accuracy are calculated.

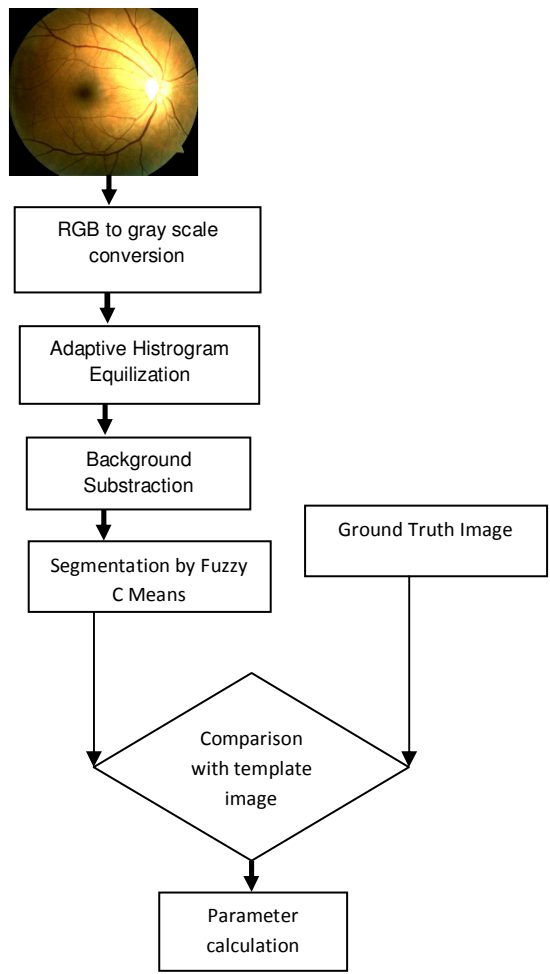

Figure 1. Proposed Method of Blood Vessel segmentation.

## 4. Explanation of the Proposed Method

In the RGB images, the green channel exhibits the best contrast between the vessels and background while the red and blue ones tend to have more noise. Therefore, we work on the gray image from green channel and the retinal blood vessels appear darker in the gray image.

Due to the acquisition process, retinal images often have a variation gray level contrast. In general, larger vessels display good contrast while the narrower ones show bad contrast. Thereby pixels attached to thick and thin vessels show the different gray level and geometrical correlation with the nearby pixels.

Normalization is performed to remove the gray-level deformation by subtracting [3] an approximate background from the original gray image. The approximate background is estimated using a 75 × 75 median filter on the original retinal image. Thereby blood vessels are brighter than the background after Normalization.

The Fuzzy C-Means algorithm includes a multiplier field, which allows the centroids for each class to vary across the image. This helps us increase prominence of every finer detail of blood vessels irrespective of thick or thin. This generates the blood vessel segmentation even with thinnest one, which was irrecoverable until then.

## 5. Results and Discussion

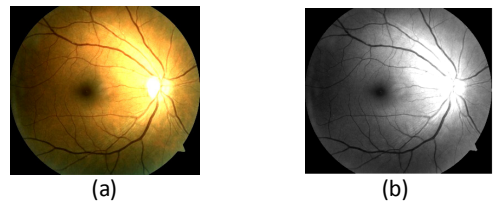

(a)                (b)

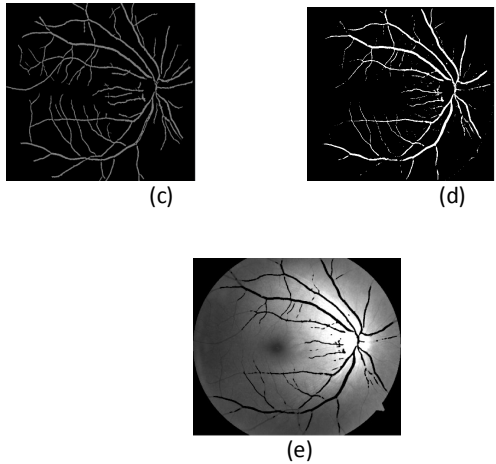

Figure 2.  (a)Original image , (b)Gray scale image , (c) Hand drawn "ground truth" ,(d) Detected Blood Vessel using proposed method,  (e) Blood vessel detected Image.

The detected results are compared with hand-drawn ground truth provided by ophthalmologists' based on nine performance measurements, namely, true positive (TP, a number of blood vessels correctly detected), false positive (FP, a number of non-blood vessels which are detected wrongly as blood vessels), false negative (FN, a number of blood vessels that are not detected), true negative (TN, a number of non-blood vessels which are correctly identified as non-blood vessels), sensitivity , specificity, positive predictive value (PPV), positive likelihood ratio (PLR ) and accuracy are calculated. This present work deals with approximately 150 data set from (The Hamilton Eye Institute Macular Edema Dataset (HEI-MED) (formerly DMED)) [9]. The specified parameters are individually calculated against each of the suitable input image and average data is given in table 1.

Sensitivity ($S_s$) = $\dfrac{TP}{TP+FN}$

Specificity ($S_P$) = $\dfrac{TN}{TN+FP}$

PPV ($P_V$) = $\dfrac{TP}{TP+FP}$

PLR ($P_R$) = $\dfrac{TP/(TP+FN)}{FP/(FP+TN)}$

Accuracy (%) = $\dfrac{TP+TN}{TP+TN+FP+FN}$

Table 1.

Average Test Case:

| Sensitivity (Ss) | Specificity ($S_P$) | PPV ($P_V$) | PLR ($P_R$) | Accuracy (%) |
|---|---|---|---|---|
| 99.62 | 54.66 | 95.08 | 219.72 | 95.03 |

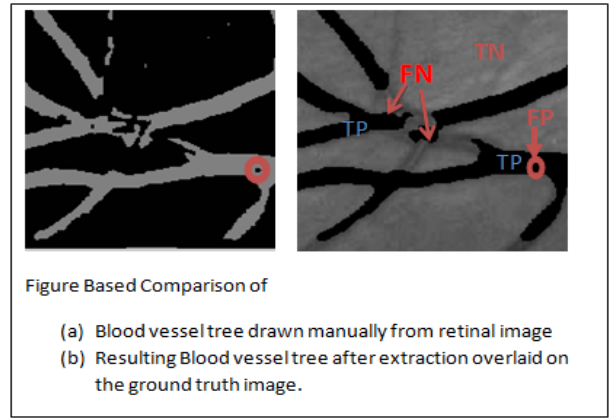

Figure Based Comparison of
(a) Blood vessel tree drawn manually from retinal image
(b) Resulting Blood vessel tree after extraction overlaid on the ground truth image.

Figure 3. FN, TN, TP, FP

## 6. Conclusion

In this present work, we deal with the hand drawn 'ground-truth' and fuzzy segmented retinal blood vessel that appears split into two parts, i.e. thick and thin vessels according to the contrast. The input images for this algorithm should be good quality in terms of sharpness, contrast, focus etc. for proper segmentation. The thick vessels are detected by adaptive local thresholding in normalized images. In this work, segmentation is done based on the FCM. The performance of the algorithm is measured against ophthalmologists' hand-drawn ground-truth. Sensitivity, specificity, PPV, PLR and sensitivity are used as the performance measurement of blood vessel detection because they combine true positive and false positive rates. A comparative study shown in Fig. 3 denotes that very little part of the vessels was not segmented properly for this further optimization can be done. The efficacy of the present work demands to be more flawless compared to standard techniques as done by the physicians from their knowledge of experiences, the present work result based using hand drawn 'ground-truth.'

**First Author**
Nilanjan Dey is an Asst. Professor at Department of Information Technology, JIS College of Engineering, Kalyani, under West Bengal University of Technology, India. He holds an M.Tech degree and a B.Tech degree in Information Technology from West Bengal University of Technology, India. The Author has 4 years of teaching experience along with 1.2 yrs of Industrial experience. His research interests are image processing, Artificial intelligence, data mining, wavelet and computation. He has 30 research papers published in National & International Journals on Image Processing & Analysis.

**Second Author**
Anamitra Bardhan Roy a final year student of Computer Science and Engineering from JIS College of Engineering, Kalyani, under West Bengal University of Technology, India, currently appointed as the Programmer Analyst Trainee in Cognizant Technology Solutions. He has 8 research papers published in National & International Journals on Image Processing & Analysis.

**Third Author**
Moumita pal is an Asst. Professor at Department of ECE, JIS College of Engineering, Kalyani, under West Bengal University of Technology, India. She holds an M.Tech degree and a B.Tech degree in ECE from West Bengal University of Technology, India. The Author has 2 years of teaching experience. Her research interests are image processing, wavelet and computation. She has 2 research papers published in International Journals on Image Processing & Analysis.

**Fourth Author**
Dr. Achintya Das was born on February 8, 1957. He received the M. Tech. and Ph.D. (Tech.) degrees in Radio Physics and Electronics from the University of Calcutta, India, in 1982 and 1996, respectively. He was an Executive of Quality Assurance with Philips India from 1982 to 1996. He is currently a Professor and Head of Electronics and Communication Engineering Dept. at Kalyani Government Engineering College, Kalyani, Nadia, West Bengal, India. His research interests include control engineering, instrumentation, Biomedical Engineering and signal processing. He is one of the reviewers of International Journal of Control, England. Special awards received by him are i). Gold Medal for securing highest position with 1st class in M.Tech. (Calcutta University) and Mohallanobis Medal Award for securing highest position with 1st. class in P.G.Dip in SQC (IAPQR, Delhi). He has 85 research papers published in National/ International journal, National/ International / world conference.